\documentclass{article}

\usepackage{microtype}
\usepackage{graphicx}
\usepackage{subcaption}
\usepackage{booktabs} % for professional tables

\usepackage{hyperref}
\usepackage{booktabs}
\usepackage{graphicx}
\usepackage{tabularx}
\usepackage{array}
\usepackage{multirow}
\usepackage{siunitx}
\usepackage{adjustbox}
\usepackage{xcolor}

\usepackage[preprint]{icml2026}

\usepackage{amsmath}
\usepackage{amssymb}
\usepackage{mathtools}
\usepackage{amsthm}
\usepackage{pgfplots}
\pgfplotsset{compat=1.18} % 设置版本（根据需要调整）
\usepackage[table]{xcolor}
\usepackage{tikz}

\usepackage[capitalize,noabbrev]{cleveref}

\theoremstyle{plain}

\theoremstyle{definition}

\theoremstyle{remark}

\usepackage[textsize=tiny]{todonotes}

\begin{document}

\twocolumn[
  \icmltitle{Unlocking Prototype Potential: An Efficient Tuning Framework for Few-Shot Class-Incremental Learning}



  \icmlsetsymbol{equal}{*}

  \begin{icmlauthorlist}
    \icmlauthor{Shengqin Jiang}{yyy}%{equal,yyy}
    \icmlauthor{Xiaoran Feng}{yyy}% {equal,yyy,comp}
    \icmlauthor{Yuankai Qi}{comp}% {comp}
    \icmlauthor{Haokui Zhang}{sch}
    \icmlauthor{Renlong Hang}{yyy}
    \icmlauthor{Qingshan Liu}{new}
    \icmlauthor{Lina Yao}{new1}
    \icmlauthor{Quan Z. Sheng}{comp}
    \icmlauthor{Ming-Hsuan Yang}{new2}
    %\icmlauthor{Firstname4 Lastname4}{sch}
    %\icmlauthor{Firstname5 Lastname5}{yyy}
    %\icmlauthor{Firstname6 Lastname6}{sch,yyy,comp}
    %\icmlauthor{Firstname7 Lastname7}{comp}
    %\icmlauthor{Firstname8 Lastname8}{sch}
    %\icmlauthor{Firstname8 Lastname8}{yyy,comp}
  \end{icmlauthorlist}

  \icmlaffiliation{yyy}{School of Computer Science, Nanjing	University of Information Science and Technology, Nanjing, 210044, China}
  \icmlaffiliation{comp}{School of Computing, Macquarie University, Sydney 2113, Australia}
  \icmlaffiliation{sch}{School of Cybersecurity, Northwestern Polytechnical University, Xi’an 710072, China}
  \icmlaffiliation{new}{School of Computer Science, Nanjing University of Posts and Telecommunications, Nanjing 210023, China}
  \icmlaffiliation{new1}{New South Wales, Sydney, NSW, Australia and CSIRO’s Data61, Sydney, NSW, Australia}
  \icmlaffiliation{new2}{University of California at Merced, Merced, CA 95343 USA}

  \icmlcorrespondingauthor{Qingshan Liu}{qsliu@nuist.edu.cn}
  \icmlcorrespondingauthor{Yuankai Qi}{qykshr@gmail.com}

  \icmlkeywords{Few-Shot Incremental-Class Learning, Prototype, Effecient Tuning}

  \vskip 0.3in
]



\printAffiliationsAndNotice{}  % no special notice (required even if empty)

\begin{abstract}
  Few-shot class-incremental learning (FSCIL) seeks to continuously learn new classes from very limited samples while preserving previously acquired knowledge. Traditional methods often utilize a frozen pre-trained feature extractor to generate static class prototypes, which suffer from the inherent representation bias of the backbone. While recent prompt-based tuning methods attempt to adapt the backbone via minimal parameter updates, given the constraint of extreme data scarcity, the model's capacity to assimilate novel information and substantively enhance its global discriminative power is inherently limited. In this paper, we propose a novel shift in perspective: freezing the feature extractor while fine-tuning the prototypes. We argue that the primary challenge in FSCIL is not feature acquisition, but rather the optimization of decision regions within a static, high-quality feature space. To this end, we introduce an efficient prototype fine-tuning framework that evolves static centroids into dynamic, learnable components. The framework employs a dual-calibration method consisting of class-specific and task-aware offsets. 
  These components function synergistically to improve the discriminative capacity of prototypes for ongoing incremental classes. Extensive results demonstrate that our method attains superior performance across multiple benchmarks while requiring minimal learnable parameters.
% our method achieves superior performance across multiple benchmarks with only a fraction of learnable parameters. 
  % our method achieves a better balance between plasticity and stability, significantly improving incremental performance with limited data.
\end{abstract}

%  To the best of our knowledge, this marks a pioneering approach to applying LLMs to few-shot entity linking tasks. OneNet is structured around three key components prompted by LLMs

\section{Introduction}
\begin{figure}[th!]
	\centering
	\includegraphics[width=0.45\textwidth]{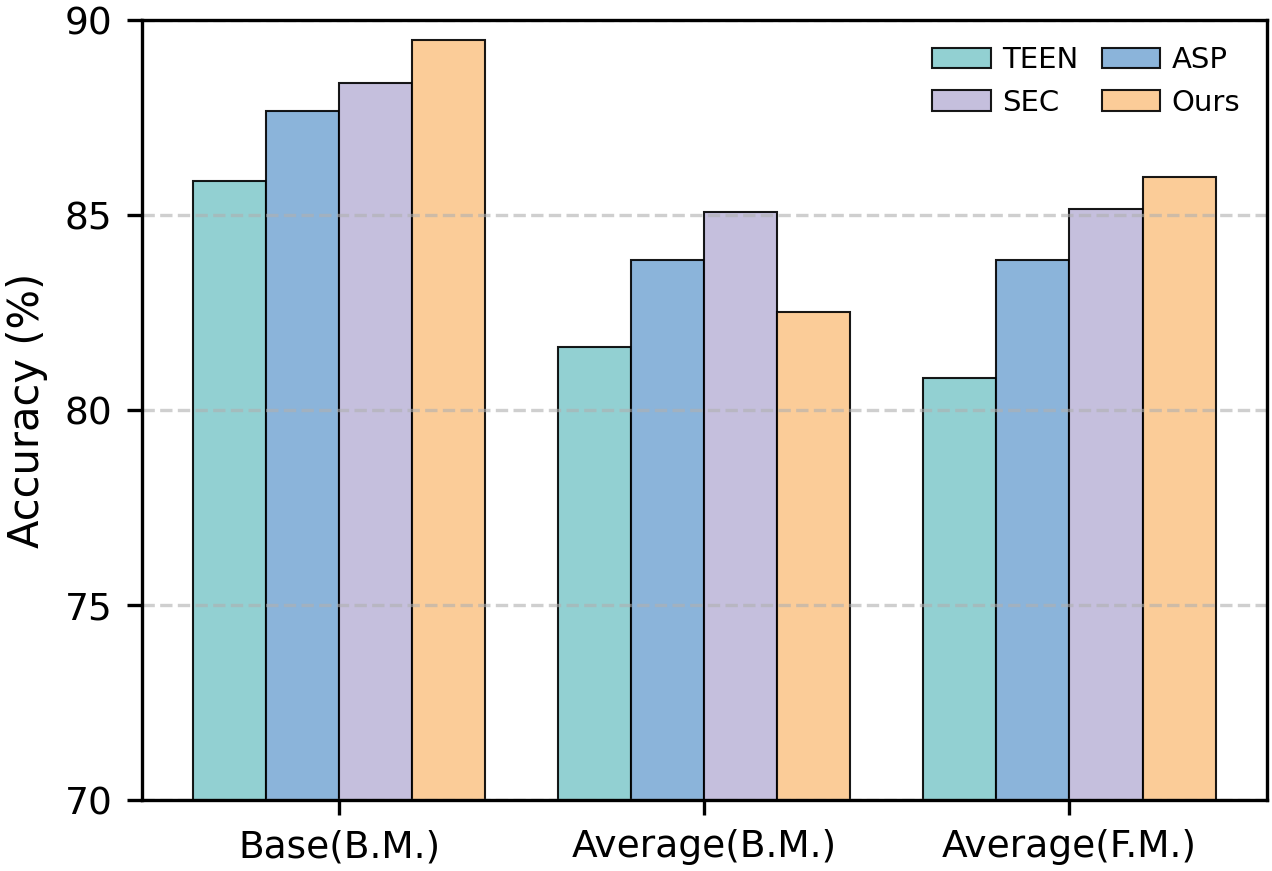}
	\caption{Performance comparison with Teen~\cite{wang2023few}, ASP~\cite{liu2024few} and SEC~\cite{liu2025sec} on  CUB-200. “Base” and “Average” indicate the accuracy of the initial stage and the mean accuracy across all incremental stages, respectively. Base Model (B.M.) denotes the model trained exclusively on base-stage data, whereas Full Model (F.M.) represents the model trained on the incremental data across all stages. For average (B.M.), the evaluation in incremental stages is conducted using prototypes generated for novel classes based on the frozen base model. Even with extremely scarce samples in the incremental phase, our model still significantly improves performance after base class learning compared to other methods.
		}
	\label{fig0}
\end{figure}

In  dynamic environments of the real world, artificial intelligence systems should, analogous to biological organisms, continuously acquire new knowledge and adapt to novel tasks while reliably preserving previously learned experiences. 
This capability, known as continual learning, is crucial for advancing models beyond static, controlled experimental settings toward practical real-world deployment. However, conventional incremental learning paradigms typically assume the availability of sufficient labeled data for each newly introduced task, an assumption that is frequently violated in practical scenarios such as medical image analysis or rare species identification, where data scarcity and annotation costs are expensive. As a result, the more challenging problem of Few-Shot Class-Incremental Learning (FSCIL) has gained increasing attention. FSCIL requires models to learn new classes sequentially from only a few labeled examples while simultaneously mitigating catastrophic forgetting of previously acquired knowledge.

In the early stage, some deep learning methods~\cite{zhu2021self,zhou2022forward} based on prototype representation have been proposed to characterize different categories. 
A typical approach is to freeze the feature extractor pre-trained on base classes and directly use the prototypes of both old and new classes as classifier weights.
The rationale behind this strategy is that freezing the feature extractor helps retain the representational capacity for old-class samples, thereby mitigating the forgetting issue, while using prototypes for incremental few-shot classes can prevent overfitting due to limited data. However, as noted in prior studies~\cite{liu2020prototype}, the feature extractor itself tends to be biased toward representing old-class samples, resulting in limited generalization ability on new classes.
This leads to significant bias in prototypes constructed from only a few new samples, which in turn restricts the model’s recognition performance on new categories.
Although subsequent work~\cite{wang2023few} has attempted to enhance the representational capacity of prototypes in new tasks through strategies such as adaptive weighting, the inherent bias of the feature extractor still limits the improvement in model performance. As shown in Fig.~\ref{fig0}, compared to the base model of TEEN, the full model after incremental learning exhibits a performance degradation across all incremental stages. This degradation can be attributed to overfitting on incremental classes with limited samples, which undermines the model’s overall discriminative capability.

%  Nevertheless

More recently, fine-tuning pre-trained models has emerged as an effective paradigm for data-scarce incremental learning tasks~\cite{liu2024few,li2025prompt}, benefiting from their strong representation learning and transfer capabilities. 
A straightforward approach is to fine-tune all model parameters; however, as observed in prior studies, the limited number of incremental samples is often insufficient to reliably optimize the large parameter space, thereby restricting performance gains. 
To mitigate these challenges, prompt-based tuning methods~\cite{liu2025sec,he2025dss,jiang2025revisiting} have gained significant traction. These approaches integrate learnable prompts into specific Transformer layers, facilitating task adaptation by optimizing only a minimal parameter subset. Consequently, they effectively preserve the generalization capabilities of the pre-trained backbone while enabling efficient downstream adaptation, yielding competitive results in few-shot scenarios. However, such performance gains during incremental stages may primarily be attributed to the inherent generalization capability of the pre-trained backbone. As illustrated in Fig.~\ref{fig0}, the full models of neither ASP~\cite{liu2024few} nor SEC~\cite{liu2025sec} demonstrate a substantial improvement in accuracy over their respective base models on incremental tasks (i.e., Average(F.M.) VS Average(B.M).
%compared to performance their base model on incremental tasks, i.e., average (B.M.). 
This suggests that while the pre-trained model retains robust generalization on unseen categories, it struggles to extract additional discriminative power from the incremental stages. Ultimately, under conditions of extreme data scarcity, the model’s capacity to assimilate novel knowledge and substantively enhance its global discriminative power remains fundamentally constrained.

% However, in incremental learning settings, newly acquired and previously learned knowledge become inevitably coupled through updates to the shared architecture. This coupling can lead to the distortion or corruption of feature representations associated with prior tasks, thereby hindering sustained performance gains. As suggested in Fig.~\ref{fig0}, SEC does not demonstrate a substantial improvement in generalization regarding novel classes when compared to the base model.

To address these challenges, we pivot from traditional backbone adaptation to a strategy of freezing the feature extractor while fine-tuning prototypes. This approach is grounded in the observation that modern pre-trained backbones already provide a highly expressive and generalizable feature space. Consequently, the primary hurdle in incremental learning shifts from feature acquisition to the optimization of decision regions within a static space. The objective becomes defining accurate, well-separated class boundaries for novel categories under low-data regimes while rigorously maintaining the integrity of previously established representations.

Guided by this insight, we focus optimization exclusively on the calibration of class prototypes. We introduce an efficient fine-tuning framework that refines class representations through two complementary learnable components: class-specific offsets, which enhance the discriminability of individual class centroids, and task-aware offsets, which model global inter-class relationships to facilitate robust task-level separation. By synergistically leveraging these offsets, our method preserves the intrinsic representational fidelity of the frozen backbone while injecting task-relevant discriminative context. The main contributions of this work are threefold:
\begin{itemize}
  \item To the best of our knowledge, we are the first to introduce an efficient prototype fine-tuning framework for FSCIL. Unlike conventional paradigms that prioritize backbone optimization, our framework evolves static prototypes into dynamic learnable components, shifting the focus toward refinement within the decision space.
  \item We introduce a dual-calibration method that augments existing prototypes through class-level and task-level offsets. It enables static prototypes to evolve into dynamic representations that characterize the feature distribution of the current task, facilitating robust adaptation in incremental settings.
  \item Extensive experiments demonstrate that our method achieves SOTA performance with minimal parameter overhead.
\end{itemize}

%  To the best of our knowledge, this marks a pioneering approach to applying LLMs to few-shot entity linking tasks. OneNet is structured around three key components prompted by LLMs

\begin{figure*}[th!]
	\centering
	\includegraphics[width=0.8\textwidth]{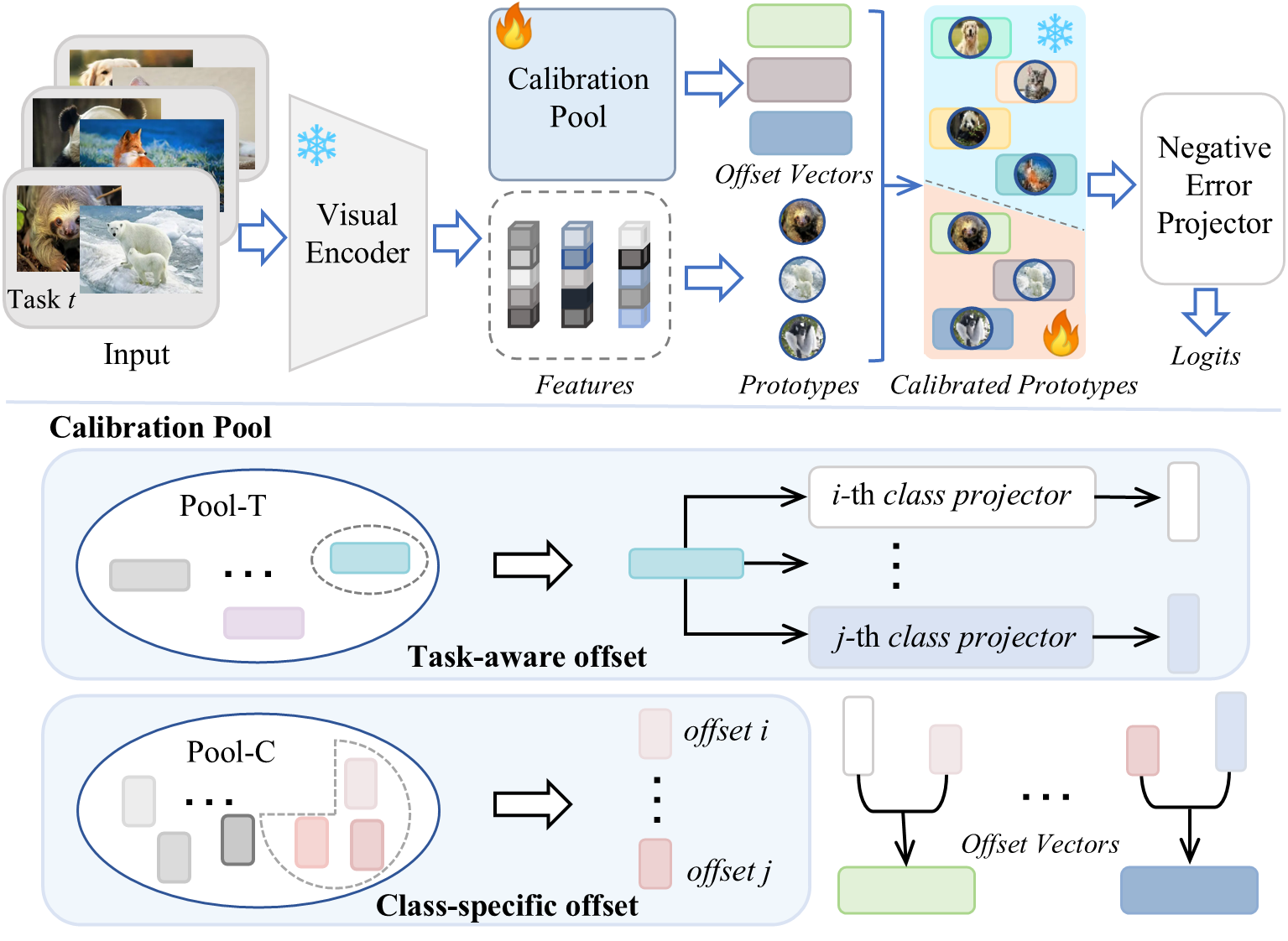}
	\caption{Overview of the proposed framework for few-shot class-incremental learning. We propose a prototype fine-tuning framework built upon a frozen pre-trained backbone. Within this architecture, each prototype is decomposed into two constituent elements for each class: a base prototype and a learnable offset. The base prototype is initialized via the global average of features extracted from the training dataset. The offset is further subdivided into a category-specific offset (sub-Sec.~\ref{CSO}) and a task-aware offset (sub-Sec.~\ref{TAO}), both of which are fine-tuned to calibrate the prototype and enhance its representational fidelity. Finally, query features are projected into the calibrated prototype space via a negative error projector (sub-Sec.~\ref{NEP}) for final prediction.
		}
	\label{net}
\end{figure*}

\section{Related Work}

\subsection{Class-Incremental Learning}
\noindent\textbf{Non-prompt-based methods.} In the context of class-incremental learning, conventional non-prompt-based methods employ diverse mechanisms to mitigate catastrophic forgetting. Replay-based approaches (e.g., iCaRL \cite{rebuffi2017icarl}, GR \cite{shin2017continual}) preserve historical knowledge by retaining a subset of previous samples or generating pseudo-samples for joint optimization with novel data. Regularization-based methods (e.g., GEM \cite{lopez2017gradient}, A-GEM \cite{chaudhry2018efficient}) impose constraints on parameter updates to ensure that optimization for new tasks does not compromise performance on established ones. Meanwhile, knowledge distillation frameworks (e.g., LwF \cite{li2017learning}, PODNet \cite{douillard2020podnet}) utilize the preceding model as a teacher to transfer semantic knowledge embedded in either prediction distributions or latent feature representations to the student model. While effective to a degree, these strategies are often hindered by practical constraints, such as the requirement for historical data buffers, the overhead of maintaining legacy model checkpoints, or computationally intensive optimization strategies.

% In class-incremental learning, non-prompt-based methods alleviate catastrophic forgetting through various mechanisms. Replay-based methods (e.g., iCaRL \cite{rebuffi2017icarl}, GR \cite{shin2017continual}) preserve old knowledge by storing samples from previous classes or generating pseudo-samples, which are jointly optimized with new data during subsequent training stages. Regularization-based methods (e.g., GEM \cite{lopez2017gradient}, A-GEM \cite{chaudhry2018efficient}) constrain the parameter update directions to prevent the optimization for new tasks from degrading performance on previously learned tasks. Knowledge distillation-based methods (e.g., LwF \cite{li2017learning}, PODNet \cite{douillard2020podnet}) treat the old model as a teacher and transfer semantic knowledge embedded in prediction distributions or feature representations to the new model.
% Although these strategies can mitigate catastrophic forgetting to some extent, they typically rely on access to historical data, the availability of old models, or the introduction of complex optimization procedures.

\noindent\textbf{Prompt-based methods.}
% To alleviate catastrophic forgetting in class incremental learning, L2P \cite{wang2022learning} proposed an incremental learning framework based on prompt pool retrieval, where the most relevant prompts were selected for each input sample to adapt to new tasks. DualPrompt \cite{wang2022dualprompt} built upon this framework by introducing task shared prompts and task specific prompts to enhance knowledge sharing across tasks. CODA-Prompt \cite{smith2023coda} further adopted an attention mechanism to weight and combine prompts, thereby optimizing the prompt selection process. S-Prompt \cite{wang2022s} leveraged the multimodal pretrained model CLIP and jointly optimized visual and textual prompts to enhance semantic alignment. Recently, CAPrompt \cite{li2025caprompt} guided inference with a frozen backbone by dynamically aggregating prompts from previously learned tasks.However, all the above methods relied on sufficient new class samples to effectively optimize prompt parameters. When the number of samples per new class was extremely limited, prompts struggled to learn stable and generalizable representations, which easily led to overfitting and consequently limited their applicability in few shot class incremental learning scenarios.
To mitigate catastrophic forgetting in class-incremental learning, L2P \cite{wang2022learning} introduced a framework based on prompt pool retrieval, selecting the most relevant prompts for each input to facilitate task adaptation. DualPrompt \cite{wang2022dualprompt} extended this architecture by decoupling prompts into task-shared and task-specific components to balance knowledge transfer and task specialization. CODA-Prompt \cite{smith2023coda} further refined this selection process by employing an attention mechanism to dynamically weight and aggregate prompts. Exploiting multimodal synergies, S-Prompt \cite{wang2022s} utilized CLIP to jointly optimize visual and textual prompts for enhanced semantic alignment, while CAPrompt \cite{li2025caprompt} guided inference through the dynamic aggregation of prompts from historical tasks. Despite their success, these methods remain data-dependent, requiring a sufficient volume of novel samples to effectively optimize prompt parameters. Under the constraints of extreme data scarcity, these learnable prompts struggle to converge on stable, generalizable representations, frequently resulting in overfitting.

\subsection{Few-Shot Class-Incremental Learning} \noindent\textbf{Prototype-based methods.}
To counter the overfitting risks prevalent in data-scarce scenarios, many few-shot methods adopt prototype-based representations~\cite{snell2017prototypical}. These approaches characterize each class by its feature centroid and utilize metric learning for sample classification.
Nevertheless, given the extreme sparsity of novel-category samples, direct point estimation of prototypes is highly susceptible to significant statistical bias, failing to represent the true underlying class distribution~\cite{tao2020few}. To address this issue, FACT \cite{zhou2022forward} allocated virtual class prototypes during pretraining to reserve embedding space for novel class prototypes. TEEN \cite{wang2023few} proposed a training free prototype calibration strategy that corrected the distributional bias of novel class prototypes by fusing them with semantically related base class prototypes. While these methods mitigate prototype bias to some extent, they often rely on backbone adaptation, which remains highly susceptible to overfitting under extreme data scarcity. This instability may trigger a representation drift that compromises previously acquired knowledge, thereby exacerbating the stability-plasticity dilemma.

% Although these methods alleviated prototype bias to some extent, they typically relied on backbone fine tuning and still struggled to avoid the erosion of previously learned knowledge under extremely limited data conditions.

\noindent\textbf{Prompt-based methods.}
% By freezing the pretrained backbone and introducing a small number of learnable prompt parameters to adapt to new tasks, prompt-based methods alleviated catastrophic forgetting while reducing the risk of overfitting under few-shot settings. 
Recently, pre-trained models have garnered significant attention in FSCIL due to their robust generalization capabilities~\cite{liu2024few,zou2024compositional}. To facilitate task adaptation while preserving the integrity of the pre-trained feature space, several approaches freeze the backbone network and introduce a minimal set of learnable parameters. Notably, prompt-based methods integrate learnable tokens into Transformer layers, enabling the model to incrementally capture evolving concepts through targeted fine-tuning of these prompts. For example, SEC-Prompt \cite{liu2025sec} achieved semantic complementarity through discriminative and non-discriminative prompts, and DSS-P~\cite{he2025dss} enabled training-free incremental learning by combining static domain prompts with dynamic instance-aware prompts. To address representation saturation caused by excessive token-level prompt concatenation, LGSP \cite{jiang2025revisiting} abandoned conventional token-based prompts and instead generated spatial prompt maps by coupling local spatial features with global frequency-domain representations, thereby alleviating feature competition. Despite significant advancements in few-shot incremental learning, the challenge of efficiently optimizing pre-trained models with extremely limited incremental data remains an open research question.

%  PCL \cite{li2025prompt} modeled prompts as a distribution over basic concepts and class–concept weights, enabling a transition from sample-level fitting to abstract concept generalization.

\section{Method}

\noindent\textbf{Overview.} In this paper, we propose an efficient prototype calibration framework, as shown in Fig.~\ref{net}. %Rather than fine-tuning the entire feature extractor, our method directly optimizes class prototypes, thereby preserving the generalization ability of the pre-trained model while enabling rapid adaptation to novel classes.
Instead of the computationally intensive fine-tuning of the entire feature extractor, our approach directly optimizes class prototypes. This strategy preserves the rich generalization of the pre-trained backbone while facilitating rapid adaptation to novel classes.
The process begins by extracting initial class prototypes using a frozen pre-trained backbone. To bolster discriminative power, we introduce class-specific and task-aware offsets that calibrate these prototypes within the feature space, ensuring a more robust and separable representation of individual categories. Finally, we employ a negative error projection metric to map query features into the calibrated prototype space, enabling efficient and robust class matching. As incremental tasks progress, these offsets are gradually accumulated to construct a calibration pool. This calibration pool is further divided into two sub-pools: a task pool (Pool-T) and a class pool (Pool-C). The core components of this architecture are detailed below.
%Specifically, we first extract initial class prototypes using a frozen pre-trained backbone. 
%To bolster discriminative power, class-specific and task-aware offsets are introduced to calibrate prototypes in the feature space, ensuring a more robust representation of individual classes. Finally, we adopt a metric based on negative error projection to map query features into the prototype space for efficient and robust class matching. We provide details of main components below.

\noindent\textbf{Problem Definition}
In the setting of FSCIL, the training dataset is defined as $\mathcal{D} = \{\mathcal{D}^0, \mathcal{D}^1, \ldots, \mathcal{D}^T\}$, where $\mathcal{D}^0$ denotes the dataset of the base task and $\mathcal{D}^t (t > 0)$ denotes the dataset of the $t$-th incremental task.
For each task, the dataset is defined as $\mathcal{D}^t = \{(x_i^t, y_i^t)\}_{i=0}^{n_t}$, where $x_i^t \in \mathbb{R}^d$ represents the feature of the $i$-th training sample and $y_i^t \in \mathcal{Y}_t$ denotes its corresponding class label. The label spaces of different tasks are disjoint $\mathcal{Y}_i \cap \mathcal{Y}_j = \varnothing$ for $i \neq j$.
In the base task, the model is trained on the dataset $\mathcal{D}^0$, which contains sufficient samples per class to learn effective representations for the base class set $\mathcal{Y}_0$. During each incremental task, the model can only access the data from the current task $\mathcal{D}^t$ for training, and historical data are not available for replay. Therefore, the model is required not only to learn newly introduced classes but also to maintain performance on previously learned classes, thereby mitigating catastrophic forgetting. After each incremental task, the model is evaluated on a test set that includes all classes encountered up to that point. Specifically, after the $t$-th incremental task, the evaluation label space is defined as $\mathcal{Y}_0 \cup \mathcal{Y}_1 \cup \cdots \cup \mathcal{Y}_t$, and the model is expected to correctly classify samples from all these classes.

% Therefore, the model is required not only to learn newly introduced classes but also to preserve the performance on previously learned classes, so as to avoid catastrophic forgetting. After completing each incremental task, the model is evaluated on a test set containing all previously seen classes. Specifically, after the $t$-th incremental task, the evaluation label space is given by $\mathcal{Y}_0 \cup \mathcal{Y}_1 \cup \cdots \cup \mathcal{Y}_t$, and the model is expected to correctly classify samples from all these classes.

\subsection{Prototype Representation}

% In incremental learning, prototypes serve as representative embeddings of classes and provide an effective mechanism to alleviate catastrophic forgetting. Inspired by ProtoNet, we adopt prototypes as the basis for classification.  

Due to the limited number of training samples, models are highly susceptible to overfitting, which in turn exacerbates catastrophic forgetting in incremental learning settings. To mitigate this issue, we adopt class prototypes as category representations~\cite{snell2017prototypical}. In the embedding space, each category is characterized by a prototype that acts as a representative anchor, encouraging samples from the same category to cluster tightly around it. Given a frozen pretrained feature extractor $f(\cdot)$, an input sample is mapped to a feature vector in the embedding space. For a class $c$, its prototype is defined as the mean feature of all support samples belonging to that class:
\begin{equation}
	{p}_c^{\text{raw}} = \frac{1}{|\mathcal{S}_c|} \sum_{x \in \mathcal{S}_c} f(x),
\label{eq:1}
\end{equation}
where $\mathcal{S}_c$ denotes the support set of class $c$.

In the base task, the availability of sufficient training data enables reliable estimation of class prototypes. Conversely, during incremental tasks, prototypes can only be inferred from only a few samples, which inevitably introduces statistical bias. As the learning process unfolds, this bias accumulates over successive tasks, substantially impairing the quality of class representations across tasks. To alleviate this issue, we propose an efficient fine-tuning strategy that calibrates prototype representations through class-level and task-level offsets, to improve their discriminability as elaborated in the subsequent sections.

%we propose an efficient fine-tuning strategy to refine prototype representations for improving their discriminability by class-specific and task-specific offsets, which will be detailed in what follows.

% In the base task, sufficient training data are available, allowing reliable estimation of class prototypes. However, in incremental tasks, prototypes are estimated from extremely limited samples, which leads to severe statistical bias. As incremental learning proceeds, prototypes of all seen classes are accumulated and used for subsequent classification. To address the bias caused by few-shot estimation, we propose a dual prompt pool based prototype enhancement strategy. Specifically, class prompts are introduced to correct prototype representations, while task prompts together with an inter class constraint promote discriminative separation among prototypes.

\subsection{Class-Specific Offset} 
\label{CSO}

Although pre-trained models exhibit strong generalization to unseen categories, their performance is often limited when directly transferred to downstream tasks. To alleviate this issue, we introduce a learnable class-specific offset for each class prototype. Acting as a lightweight and independent regulator, the offset vector enables fine-grained adaptation of class representations while preserving the stability of the core features learned from pre-trained task.

Specifically, for each incremental task, we first compute the original class prototypes ${p}_c^{\text{raw}} \in \mathbb{R}^{d_f}$ using a frozen feature extractor.
Conditioned on these prototypes, we initialize a learnable offset vector ${p}_c^{\text{class}}$ for each category by sampling from a scaled Gaussian distribution, i.e., ${p}_c^{\text{class}} \sim \alpha \cdot \mathcal{N}(\mathbf{0}, \mathbf{I})$, where $\alpha$ controls the magnitude of initialization values. This design ensures that, in the early stages of training, the effective category representation is dominated by the original prototype, with only a minimal contribution from the offset vector. As a result, optimization starts from a stable initialization, reducing training oscillations and facilitating smooth and reliable convergence. Upon the arrival of a new incremental task, we apply a consistent procedure to the novel classes: initializing their original class prototypes alongside their corresponding offset vectors.

\subsection{Task-Aware Offset} 
\label{TAO}

Class-specific offsets primarily refine individual category representations but overlook class relationships across tasks. To this end, we introduce a task-aware offset that models task-specific data distribution biases and enables targeted calibration. This offset explicitly characterizes the deviation between the data distribution of the current task and the general pre-trained distribution. Consequently, the model can consistently adjust the prototypes of all categories in a task-aware manner, guided by the global characteristics of the task. This design facilitates more accurate adaptation to task-specific data properties and leads to more well-defined classification boundaries.

To be more specific, for each task $t$, we introduce a learnable task offset vector
${p}_t^{\text{task}} \in \mathbb{R}^{d_t}$,
where $d_t$ denotes the dimensionality of the task offset.
The offset is initialized to zero and shared among all classes within the same task. For each class $c \in \mathcal{C}_t$, the task offset is fed into a dependent class projector to generate an offset vector ${o}^{\text{task}}_c$. Such a design allows the shared task offset to generate distinct offset representations for different classes. Each projector is implemented as a two-layer multilayer perceptron (MLP) with a hidden dimension $d_h$.  Formally, it is formulated as:
\begin{equation}
	{o}^{\text{task}}_c = \mathrm{MLP}_c({p}_t^{\text{task}}).
	\label{eq:task_offset}
\end{equation}

To further enhance class separability, we introduce an inter-class discrimination loss $L_{\text{inter}}$ to regularize the calibrated prototypes, which is described in the next section. The final calibrated prototypes for class $c$ are defined as:
\begin{equation}
	{p}_c
	=
	{p}_c^{\text{raw}}
	+
	{p}_c^{\text{class}}
	+
	{o}^{\text{task}}_c.
	\label{eq:final_prototype}
\end{equation}

After completing each task, we freeze these calibrated prototypes ${p}_c$.

% During training, the task prompts, class prompts, and parameters of the class-specific MLPs are jointly optimized via backpropagation. After the completion of each task, these parameters are frozen to prevent catastrophic forgetting in subsequent tasks.

% --------------------------------------------------

\subsection{Negative Error Projector} 
\label{NEP}
%%%%%%%%%%added Matching network  references

Given prototypes, existing methods~\cite{snell2017prototypical,liu2024few} typically employ distance metrics such as Euclidean distance~\cite{cheraghian2021semantic} or cosine similarity~\cite{vinyals2016matching} to assign a query feature to the category corresponding to its most similar prototype. These methods treat each prototype as an isolated point in the feature space, overlooking the potential inter‑relationships among prototypes. Especially in the few‑shot incremental learning scenario, prototypes derived from a very limited number of samples are often unstable and may not reliably represent the entire category. This further complicates the accurate discrimination of query features.

% This further complicates the discrimination of query features.

To address this issue, we adopt a negative error projector for classification. Specifically, the query feature is formulated as a linear combination of all class prototypes, where the optimal combination coefficients are obtained by minimizing the reconstruction error \cite{zhang2011sparse}. 
The category of the query is then determined based on the residual discrepancy between the reconstructed feature and the original query feature. 
This formulation explicitly captures fine-grained linear relationships between the query feature and all prototypes, thereby modeling continuous semantic information that reflects the degree to which a sample belongs to each category. Consequently, it enables more refined and principled classification decisions.
%In few-shot class-incremental learning, prototype representations may remain biased due to limited samples, making distance-based classification unreliable. To further enhance robustness, we introduce a Negative Error Projector that performs classification based on reconstruction residuals with respect to enhanced prototypes.

Let $K \in \mathbb{R}^{C \times d_f}$ denote the matrix of enhanced prototypes after task  $t$, where $C$ is the number of observed classes, and let $\mathbf{f} \in \mathbb{R}^{d_f}$ denote an input feature. We obtain the reconstruction coefficient vector $\boldsymbol{\rho} \in \mathbb{R}^{C}$ by solving the following regularized least-squares problem:

\begin{equation}
	\min_{\boldsymbol{\rho}}
	\;
	\left\lVert f(x) - K^\top \boldsymbol{\rho} \right\rVert_2^2
	+
	\lambda_{\text{reg}}
	\left\lVert \boldsymbol{\rho} \right\rVert_2^2,
	\label{eq:lsq}
\end{equation}
where $\lambda_{\text{reg}}$ is a regularization parameter. The closed-form solution is $\boldsymbol{\rho}^{*} =(KK^\top + \lambda_{\text{reg}} \mathbf{I})^{-1}Kf(x)$. The normalized reconstruction residual is then computed for each class $i$ as follows:
\begin{equation}
	R_i(\mathbf{f})
	=
	\frac{\left\lVert f(x) - \boldsymbol{\rho}^{*}_i K_i \right\rVert_2}
	{\left\lVert \boldsymbol{\rho}^{*}_i \right\rVert_2 + \epsilon},
	\label{eq:residual}
\end{equation}
where $\epsilon$ is a small constant. 

% By introducing subspace-based reconstruction and competitive constraints among global prototypes, the Negative Error Projector effectively corrects prototype bias and enhances discriminability between visually similar classes without increasing the training burden.

\noindent\textbf{Optimization.} Prior prototype-based networks typically treat a class prototype as a rigid statistical anchor, namely, the mean vector of support features within a latent space. In these frameworks, prototypes are updated only indirectly through global parameter optimization, requiring a full recomputation whenever the feature extractor changes.

Our proposed method decouples the prototype into a dual-component architecture: a base statistical prototype and a learnable offset vector. By directly optimizing this offset, we transition the class representation from a static, one-time calculation into a dynamic, task-adaptive embedding. This mechanism allows for accurate calibration through a joint objective of the cross-entropy classification loss $L_{\text{CE}}$ and inter-class discrimination loss $L_{\text{inter}}$. 
%Given a sample feature $f({x}_i) \in \mathbb{R}^{d_f}$ with label $y_i$, we define $\mathcal{P}_{\text{neg}}$ as the set of enhanced prototypes belonging to all classes other than $y_i$, including both current and previously learned classes. 
Specifically, for a given sample $x_i$ with the label $y_i$, we define $\mathcal{P}_{\text{neg}}$ as the set of calibrated prototypes excluding the prototype corresponding to $y_i$. This set encompasses all class representations from both the current stage and previously encountered classes, effectively serving as the set of negative class anchors for the current sample.
The inter-class loss for sample $x_i$ is defined as:
\begin{equation}
	L_{\text{inter}}(x_i)
	= \frac{1}{|\mathcal{P} _{\mathrm{neg}}|}\frac{1}{\sum_{p_j\in \mathcal{P} _{\mathrm{neg}}}{\left\| f(x_i)-p_j \right\|}+\epsilon},
	\label{eq:inter_loss}
\end{equation}
where $\epsilon$ is a small constant.

The overall training objective is formulated as:
\begin{equation}
	L = L_{\text{CE}} + \lambda_{\text{inter}} L_{\text{inter}},
	\label{eq:total_loss}
\end{equation}
where $\lambda_{\text{inter}}$ is a hyper-parameter. 

\begin{table*}[t]
	\centering
	\caption{Performance comparison against SOTA methods on CUB-200.  We re-implement TEEN and Comp, replacing their original backbones with a pre-trained ViT integrated via an adapter. EPT and EPT$^*$ refer to our method implemented with DINOv2 and DINOv3 backbones, respectively. Params denotes the learnable parameters.  \textbf{Avg}: average of all stages.}
	\label{tab:cub_results}
	
	\setlength{\tabcolsep}{2.2pt}
	\renewcommand{\arraystretch}{1.0}
	
	\begin{tabularx}{\textwidth}{l c *{11}{>{\centering\arraybackslash}X} c}
		\toprule
		\multirow{2}{*}{\centering Method} &
		\multirow{2}{*}{\centering Params (M)} &
		\multicolumn{11}{c}{Acc. in each stage (\%)} &
		\multirow{2}{*}{\centering Avg} \\
		\cmidrule(lr){3-13}
		&  & S0 & S1 & S2 & S3 & S4 & S5 & S6 & S7 & S8 & S9 & S10 &  \\
		\midrule
		CEC~\cite{zhang2021few}        & 1.34   & 84.8   & 82.5 & 81.4 & 78.5 & 79.3 & 77.8 & 77.4 & 77.6 & 77.2 & 76.9 & 76.8 & 79.10 \\
		TEEN~\cite{wang2023few}       & 1.34   & 85.87   & 84.56 & 83.51 & 80.82 & 81.86 & 79.64 & 79.53 & 79.23 & 78.49 & 78.20 & 78.43 & 80.92 \\
        MambaF~\cite{li2024mamba}  & 20.62 & 80.90 & 76.26 & 72.97 & 70.14 & 67.83 & 65.74 & 65.43 & 64.12 & 62.31 & 62.12 & 61.65 & 68.13 \\
		ASP~\cite{liu2024few}            & 2.08  & 87.67 & 86.41   & 85.83 & 84.01 & 83.99 & 82.41 & 82.25 & 82.61   & 82.04 & 82.45   & 82.64 & 83.84 \\
		Yourself~\cite{tang2024rethinking}       & 12.7  & 83.4 & 77.0   & 75.3 & 72.2 & 69.0   & 66.8 & 66.0   & 65.6 & 64.1 & 64.5 & 63.6 & 69.85 \\
		NTK~\cite{liu2024ntk}             & 50.0    & 84.0   & 81.8 & 80.1 & 77.4 & 76.6 & 74.6 & 74.6 & 74.4 & 74.0   & 73.9 & 73.6 & 76.81 \\
		Comp~\cite{zou2024compositional}      & 1.5 & \underline{90.54} & 85.97 & 85.42 & 84.66 & 83.71 & 81.83 & 81.56 & 81.51 & 81.02 & 81.27 & 81.58 & 83.55 \\
		SEC~\cite{liu2025sec}            & 5.64  & 88.37 & 87.02 & \underline{86.71} & \underline{85.30} & \underline{85.38} & 83.66 & 83.66 & \underline{84.18} & \underline{83.98} & \underline{84.18} & \underline{84.48} & 85.17 \\
		PCL~\cite{li2025prompt}          & --    & 85.23 & 81.86 & 80.05 & 79.38 & 75.85 & 74.60 & 73.85 & 71.42 & 69.59 & 69.06 & 68.31 & 75.38 \\
		PA~\cite{liu2025prototype}              & --    & 78.69 & 75.59 & 72.71 & 68.71 & 68.37 & 65.77 & 64.75 & 63.59 & 62.76 & 62.02 & 61.19 & 67.65 \\
		DSS-P~\cite{he2025dss}      & --    &\textbf{90.95} & \underline{87.64} & 86.43 & 85.12 & 85.32 & \underline{83.99} & \underline{83.59} & 83.63 & 83.25 & 83.62 & 84.22 & \underline{85.25} \\
		PET~\cite{liu2025pet}         & --    & 82.30 & 79.23 & 77.61 & 72.40 & 72.51 & 71.29 & 70.65 & 69.15 & 68.34 & 66.85 & 67.12 & 72.49 \\
		LGSP~\cite{jiang2025revisiting}    & \textbf{0.21}  & 85.72 & 84.31 & 83.21 & 81.33 & 81.8 & 80.33 & 79.89 & 80.09 & 79.18 & 79.74 & 79.72 & 81.39 \\
		\midrule
		\rowcolor{blue!5}
		\textbf{EPT}              & \underline{0.22}   & 89.49 & \textbf{87.84} & \textbf{87.43} & \textbf{86.34} & \textbf{86.07} & \textbf{84.39} & \textbf{84.54} & \textbf{84.97} & \textbf{84.74} & \textbf{84.76} & \textbf{85.10} & \textbf{85.97} \\
		\rowcolor{blue!5}
		\textbf{EPT$^*$}            & 0.28   & 90.32 & \textbf{88.92} & \textbf{87.84} & \textbf{86.90} & \textbf{86.39} & \textbf{85.76} & \textbf{85.43} & \textbf{86.13} & \textbf{86.01} & \textbf{86.49} & \textbf{86.48} & \textbf{86.97} \\
		\bottomrule
	\end{tabularx}
\end{table*}

\section{Experiments}

\subsection{Experiment Setup}

\noindent\textbf{Datasets.} We evaluate our method across four benchmark datasets: CUB-200, ImageNet-R, ImageNet-A, and VTAB. For the first three, each comprising 200 classes, we reserve 100 classes for the base stage and distribute the remaining 100 classes across ten 10-way 5-shot incremental tasks, totaling 11 stages. For VTAB, which contains 50 classes, the base stage covers 14 classes, followed by nine 4-way 5-shot incremental tasks for a total of 10 stages.

\noindent\textbf{Training Details.} Across all datasets, the model is trained for 100 epochs during the base stage and 60 epochs per incremental stage, using a batch size of 64. Hyper-parameters are configured with $\lambda_{\text{reg}}=0.3$ for the negative error projector,  $\lambda_{\text{inter}}=0.1$ for the inter-class discrimination loss, and a scaling factor of $\alpha=0.001$. The hidden dimension of MLPs is set to $d_h=4$  for CUB-200, ImageNet-A, and VTAB, while a higher dimension of $d_h=8$ is utilized for ImageNet-R. Unless otherwise specified, all other hyperparameters remain consistent across all experimental settings. We will publicly release our code upon the acceptance of our paper.

% For all datasets, the model is trained for 100 epochs in the base stage and 60 epochs in each incremental stage, with a batch size of 64. We set $\lambda_{\text{reg}}=0.3$ for the Negative Error Projector and $\lambda_{\text{inter}}=0.1$ for the inter-class loss. the scale $\alpha$ is set to 0.001. The hidden dimension of MLPs is set to $D_h=4$ for CUB-200, ImageNet-A, and VTAB. For ImageNet-R, it is increased to $D_h=8$. All remaining hyperparameters are kept consistent across datasets unless otherwise specified.

\noindent\textbf{Evaluation Protocol.}  Following established benchmarks, we adopt the widely used Top-1 accuracy, which is denoted as $A_i$ for the $i$-th stage. The overall performance is evaluated using the average accuracy (Avg), defined as the arithmetic mean of accuracies across all stages. 
% All reported results signify Top-1 accuracy.

% Following previous works, we denote the Top-1 classification accuracy after the $i$-th stage as $A_i$. The overall performance is reported by the average accuracy (Avg), which is computed as the mean of accuracies across all stages. All results are reported in terms of Top-1 accuracy.

\subsection{Comparison With State-of-the-Art Methods}

To validate the effectiveness of the proposed method, we conduct a comprehensive comparison against SOTA approaches across multiple datasets. For simplicity, we term our method EPT, short for Efficient Prototype Tuning. Before presenting the results, we introduce two variants of our method: EPT and EPT$^*$, which utilize DINOv2 and DINOv3 as their respective backbone networks. The experimental results on CUB-200 are provided in Table~\ref{tab:cub_results}. % As shown, in the incremental stages, both of our proposed solutions consistently outperform the second-best method, DSS-P. Specifically, EPT achieves an accuracy of 85.10\% in the final stage. Furthermore, EPT$^*$ significantly surpasses EPT across all incremental tasks, yielding a 1.0\% improvement in the final average incremental accuracy. It is worth noting that our methods are highly parameter-efficient, requiring only 0.22M and 0.28M learnable parameters to be tuned, respectively.
As illustrated, both of our proposed solutions consistently outperform the second-best method, DSS-P~\cite{he2025dss}, across all incremental stages. Specifically, EPT achieves a final-stage accuracy of 85.10\%. Furthermore, EPT$^*$ significantly surpasses EPT throughout the incremental process, yielding a 1.0\% improvement in final average incremental accuracy. Notably, our methods maintain high parameter efficiency, requiring only 0.22M and 0.28M learnable parameters, respectively. We notice that the recent method LGSP~\cite{jiang2025revisiting} optimizes fewer learnable parameters than our method, with only 0.21M, but its accuracy in the final stage is 6.2\% lower than ours, and its average accuracy is 5.3\% lower. % Moreover, we present performance comparisons across ImageNet-R, VTAB, and ImageNet-A, as shown in  Fig.~\ref{fig:comparison.png}.  Our method achieves superior performance, where EPT$^*$, leveraging the DINOv3 backbone, yields obvious performance gains, with particularly significant improvements observed on ImageNet-R and ImageNet-A.

\begin{figure*}[t]
	\centering
	\includegraphics[width=1\textwidth]{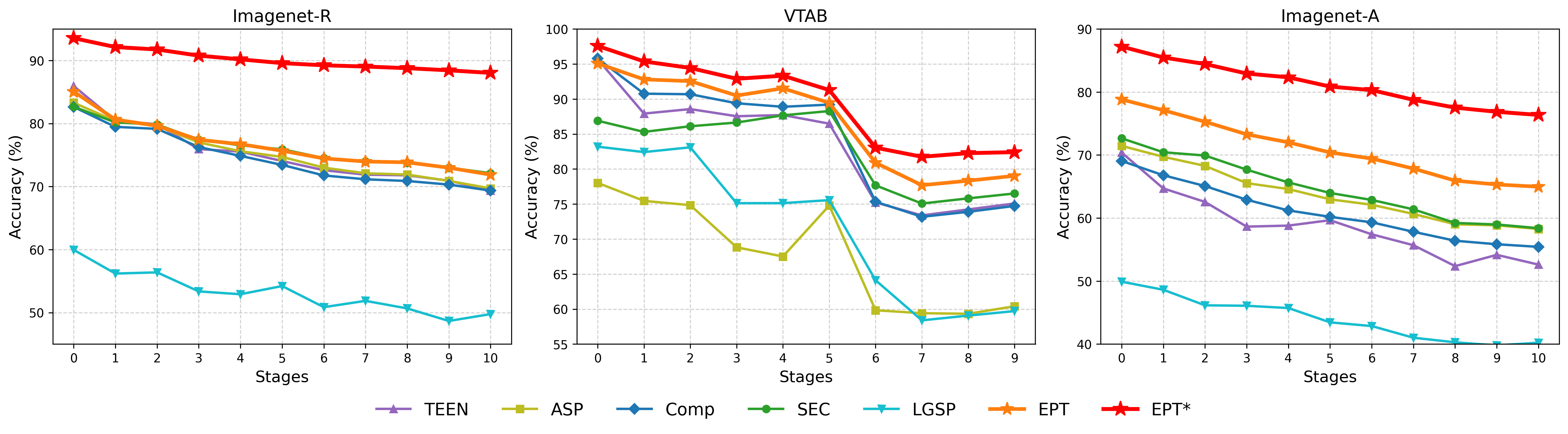}
	\caption{Performance comparison against SOTA methods on ImageNet-R, VTAB, and ImageNet-A.}
	\label{fig:comparison.png}
\end{figure*}

In addition, we provide comprehensive performance comparisons across ImageNet-R, VTAB, and ImageNet-A, as illustrated in Fig.~\ref{fig:comparison.png}. Our proposed method consistently achieves superior results. Notably, EPT$^*$, leveraging the DINOv3 backbone, demonstrates substantial performance gains, with particularly marked improvements observed on the challenging ImageNet-R and ImageNet-A datasets. These results underscore the robustness and scalability of our model in data-constrained incremental scenarios.

\begin{table*}[t]
	\centering
	\caption{Ablation studies on the key components. NEP, CS Offset, and  TA Offset denote the negative error projector, class-specific offset, and task-aware offset, respectively.}
	\label{tab:results}
	\setlength{\tabcolsep}{4pt} % 缩小列间距，避免过宽
	\begin{tabular}{@{}ccccccccccccccc@{}}
		\toprule
		\multirow{2}{*}{\centering NEP} &
		\multirow{2}{*}{\centering CS Offset} &
		\multirow{2}{*}{\centering TA Offset} &
		\multicolumn{11}{c}{Acc. in each stage (\%)} &
		\multirow{2}{*}{\centering Avg} \\
		\cmidrule(lr){4-14} % ← 只在 stage 列（第4到14列）画线
		&  &  & S0 & S1 & S2 & S3 & S4 & S5 & S6 & S7 & S8 & S9 & S10 &  \\
		\midrule
		\checkmark &  &  & 85.78 & 84.31 & 83.64 & 83.12 & 83.06 & 81.94 & 82.16 & 82.71 & 82.53 & 82.83 & 83.31 & 83.22 \\
		\checkmark & \checkmark &  & 89.07 & 87.75 & 87.17 & 86.26 & 85.97 & 84.09 & 83.96 & 84.30 & 84.13 & 84.54 & 84.46 & 85.61 \\
		\checkmark &  & \checkmark & 88.82 & 87.08 & 86.53 & 85.43 & 84.8 & 83.37 & 83.33 & 83.5 & 83.15 & 83.41 & 83.37 & 84.79 \\
		\checkmark & \checkmark & \checkmark & 89.49 & 87.84 & 87.43 & 86.34 & 86.07 & 84.39 & 84.54 & 84.97 & 84.74 & 84.76 & 85.10 & 85.97 \\
		\bottomrule
	\end{tabular}
\end{table*}

\subsection{Ablation Study}

In this subsection, we conduct ablation studies to evaluate the effectiveness of each proposed strategy on CUB-200. Unless otherwise specified, all experiments utilize DINOv2 as feature extractor. 

\noindent\textbf{Analysis of key components.} Table \ref{tab:results} summarizes the ablation studies for our model’s key components, detailing the accuracy at each stage from S0 to S10. Each module is shown to contribute positively to the overall performance. Specifically, the individual integration of Class-Specific (CS) and Task-Aware (TA) offsets yields measurable gains in model performance. While the CS offset facilitates fine-grained, independent parameter adjustments for each class prototype, the TA offset explicitly models the inter-task relationships. Our experimental results demonstrate that the simultaneous application of both offsets further enhances performance, confirming that these modules are functionally complementary. These results validate the effectiveness of the proposed approach.

\begin{figure}[t]
	\centering
	\includegraphics[width=0.5\textwidth]{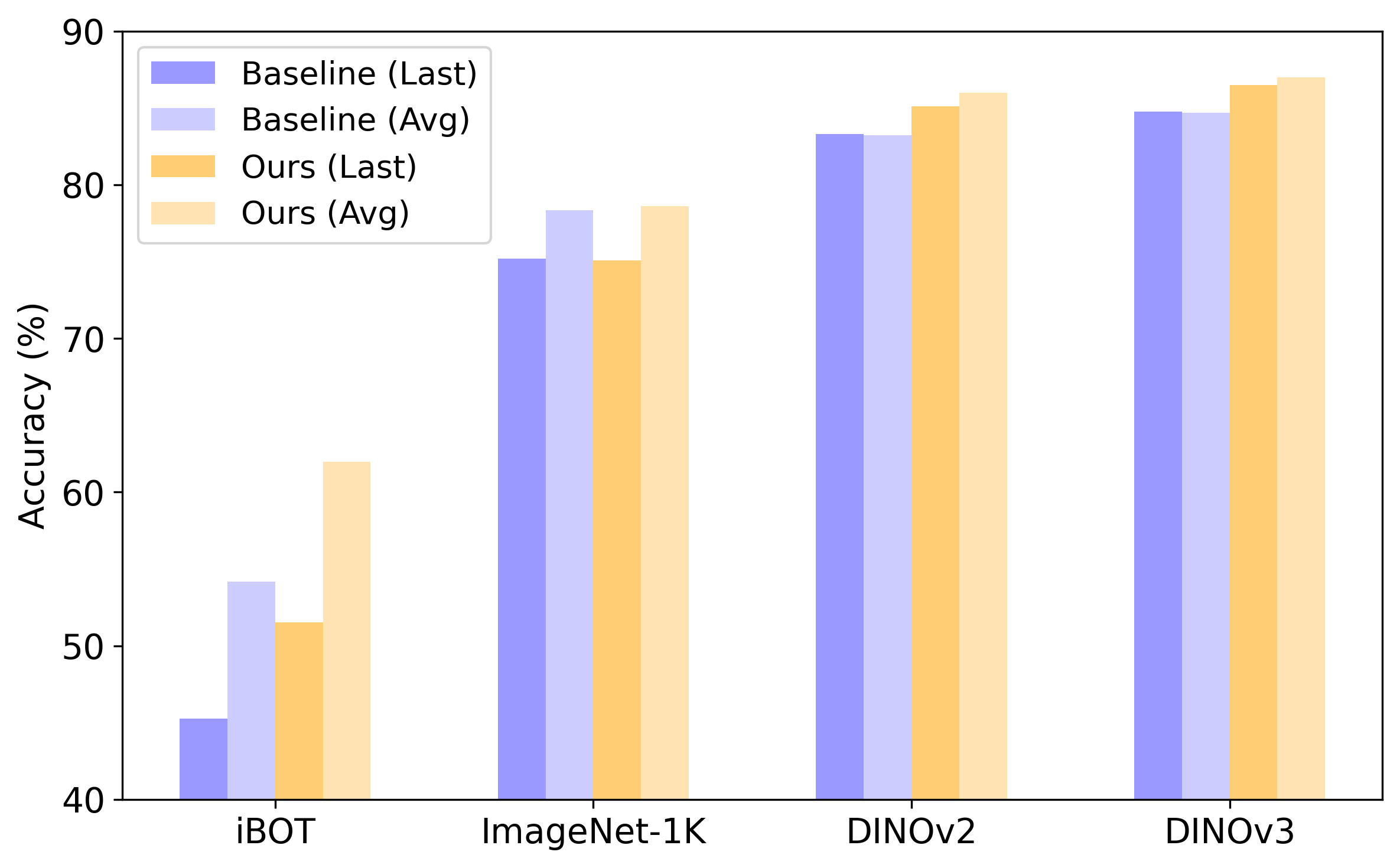}
	\caption{Performance comparison between our proposed method and the baseline across various pre-trained backbones. \textit{Last} denotes accuracy on the final task, and \textit{Avg} denotes average accuracy across all tasks.}
	\label{fig:backbone}
\end{figure}

\noindent\textbf{Analysis of backbones.} % Although our method yields slightly lower accuracy than the baseline in the final stage of incremental learning on ViT, it consistently outperforms the baseline across most stages, with particularly pronounced gains in the early stages. More importantly, our method surpasses the baseline in terms of final average accuracy, which validates the stability and adaptability of our method in long-term incremental learning tasks. Furthermore, it is worth noting that our method shows significant improvement in the DINO model, especially in the later incremental tasks. By effectively preventing catastrophic forgetting and enhancing task adaptability, EPT method maintains excellent performance across all stages. This demonstrates that our approach has stronger generalization ability in complex incremental learning tasks. In particular, in the DINO model, the performance across all stages has been significantly enhanced, further proving the efficiency and robustness of our method. The experimental results are shown in Figure \ref{fig:backbone}.
In our framework, we freeze the pre-trained backbone and facilitate incremental learning solely by fine-tuning prototypes. This strategy relies on the assumption that the backbone possesses robust generalization capabilities. To evaluate the influence of various pre-training paradigms on incremental performance, we conduct a comparative analysis using four representative models: iBOT, ImageNet-1K, DINOv2, and DINOv3. Notably, ImageNet-1K represents a supervised baseline, whereas the others are derived from self-supervised pre-training.

As illustrated in Fig.~\ref{fig:backbone},  compared to our baseline, our method can significantly improve its performance across different pre-training backbones. Here, the baseline refers to our method excluding both CS Offset and TA Offset. Furthermore, the results indicate that DINOv2 and DINOv3 achieve superior results, while iBOT yields the lowest performance. This performance gap is likely attributable to the massive datasets used to train the DINO series, which significantly bolster their representational capacity and cross-domain generalization. While the supervised ImageNet-1K model provides a competent baseline, its ability to generalize to novel incremental tasks remains markedly inferior to that of the large-scale self-supervised models. Ultimately, these findings confirm that backbones pre-trained on extensive, diverse data provide the most stable and reliable feature foundations, which are critical for effective prototype-based incremental learning.

\begin{figure}[t]
	\centering
	\includegraphics[width=0.5\textwidth]{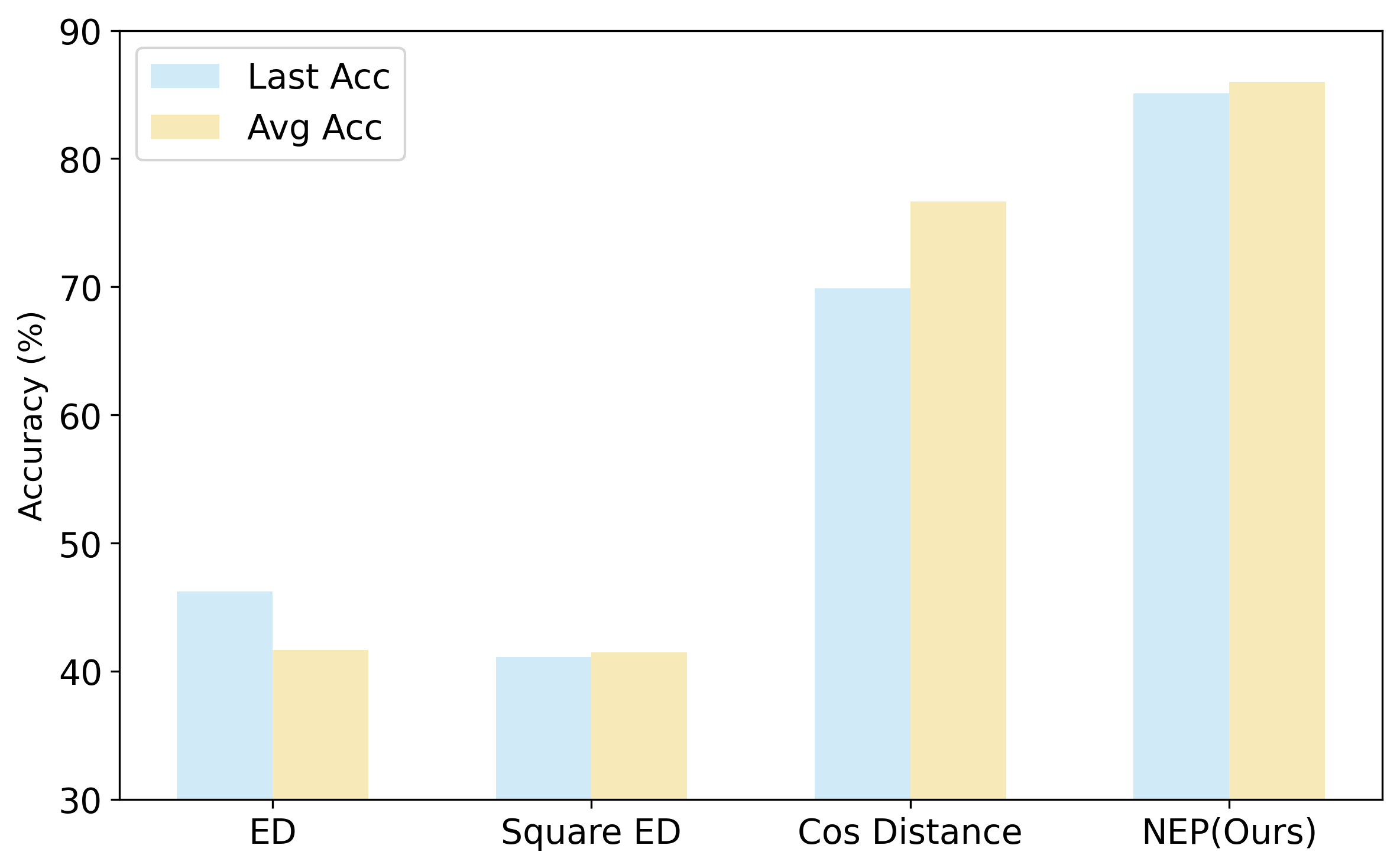}
	\caption{Performance comparison of different distance metrics for prototype-based classification. ED and NEP denote Euclidean distance and negative error projector, respectively. The plot shows the accuracy for the final task (\textit{Last Acc}) and the average accuracy (\textit{Avg Acc}) across all tasks.}
	\label{fig:metric_comparison}
\end{figure}

\noindent\textbf{Analysis of distance metrics.} % ~\cite{snell2017prototypical,vinyals2016matching}
Unlike traditional classification paradigms, prototype-based methods fundamentally rely on distance metrics to determine class membership. To evaluate the robustness of our approach, we conduct a comparative analysis across several standard metrics, including Euclidean distance~\cite{cheraghian2021semantic}, squared Euclidean distance~\cite{snell2017prototypical}, and cosine distance~\cite{vinyals2016matching, wang2023few, liu2024few}. The results are presented in Fig.~\ref{fig:metric_comparison}. As illustrated, our method consistently achieves superior performance across all metrics. Traditional approaches typically treat prototypes as isolated points within the feature space, thereby neglecting the intrinsic structural correlations between classes. In contrast, the negative error projector explicitly models these linear correlations to refine the model's offset vectors. By leveraging this relational information, we significantly enhance the discriminative power of the prototypes.

%In this feature distribution plot, we show the class prototypes and sample feature points computed using the DINOv2 as feature extractor. As shown in Figure \ref{fig:features.png}, the plot includes 100 classes, 50 base classes, and 5 randomly selected classes for each incremental task. The DINOv2 pretrained model is already able to effectively distinguish between classes, with a clear separation between base class and incremental task samples in the feature space, demonstrating its strong class distinction ability. Building on this excellent feature extractor, our method further optimizes it, enhancing the distinction between features and making the distribution of each class clearer, ensuring that the boundaries between classes are more defined in incremental learning. This optimization greatly improves the model’s performance in incremental tasks, reduces catastrophic forgetting, and enables better adaptation to new classes.

%  Few-shot class-incremental learning seeks to continuously learn new classes from very limited samples while preserving previously acquired knowledge. This task is fundamentally hindered by catastrophic forgetting and overfitting. Existing methods typically depend on empirically defined statistical relationships between base and novel class prototypes, which constrain their adaptability and generalization. 

\section{Conclusion}

This work studies the core challenges of FSCIL, namely, catastrophic forgetting and overfitting, by rethinking the conventional fine-tuning paradigm. Given the high risk of updating a pre-trained backbone with scarce incremental data, we freeze the feature extractor to fully exploit its well-structured latent space and instead concentrate optimization on class prototypes. To this end, we introduce dynamic, learnable prototypes refined through a dual-calibration method that incorporates both class-specific and task-aware offsets. The former offsets improve the discriminability of individual class centroids, whereas the latter offsets capture global inter-class relationships, enabling robust separation at the task level. Moreover, we employ a negative error projector to establish and optimize the relationship between the features of the query sample and the calibrated prototypes, thereby achieving the final classification decision. Extensive experiments demonstrate that the proposed method achieves SOTA performance on multiple benchmarks with negligible parameter overhead.

\section*{Impact Statement}

This paper presents work whose goal is to advance the field of machine learning. There are many potential societal consequences of our work, none of which we feel must be specifically highlighted here.

\nocite{langley00}

\bibliography{cvio}
\bibliographystyle{icml2026}

%%%%%%%%%%%%%%%%%%%%%%%%%%%%%%%%%%%%%%%%%%%%%%%%%%%%%%%%%%%%%%%%%%%%%%%%%%%%%%%
%%%%%%%%%%%%%%%%%%%%%%%%%%%%%%%%%%%%%%%%%%%%%%%%%%%%%%%%%%%%%%%%%%%%%%%%%%%%%%%
% APPENDIX
%%%%%%%%%%%%%%%%%%%%%%%%%%%%%%%%%%%%%%%%%%%%%%%%%%%%%%%%%%%%%%%%%%%%%%%%%%%%%%%
%%%%%%%%%%%%%%%%%%%%%%%%%%%%%%%%%%%%%%%%%%%%%%%%%%%%%%%%%%%%%%%%%%%%%%%%%%%%%%%
\newpage
\appendix
\onecolumn
\section{Appendix}

\begin{figure}[t]
	\centering
	\includegraphics[width=0.5\linewidth]{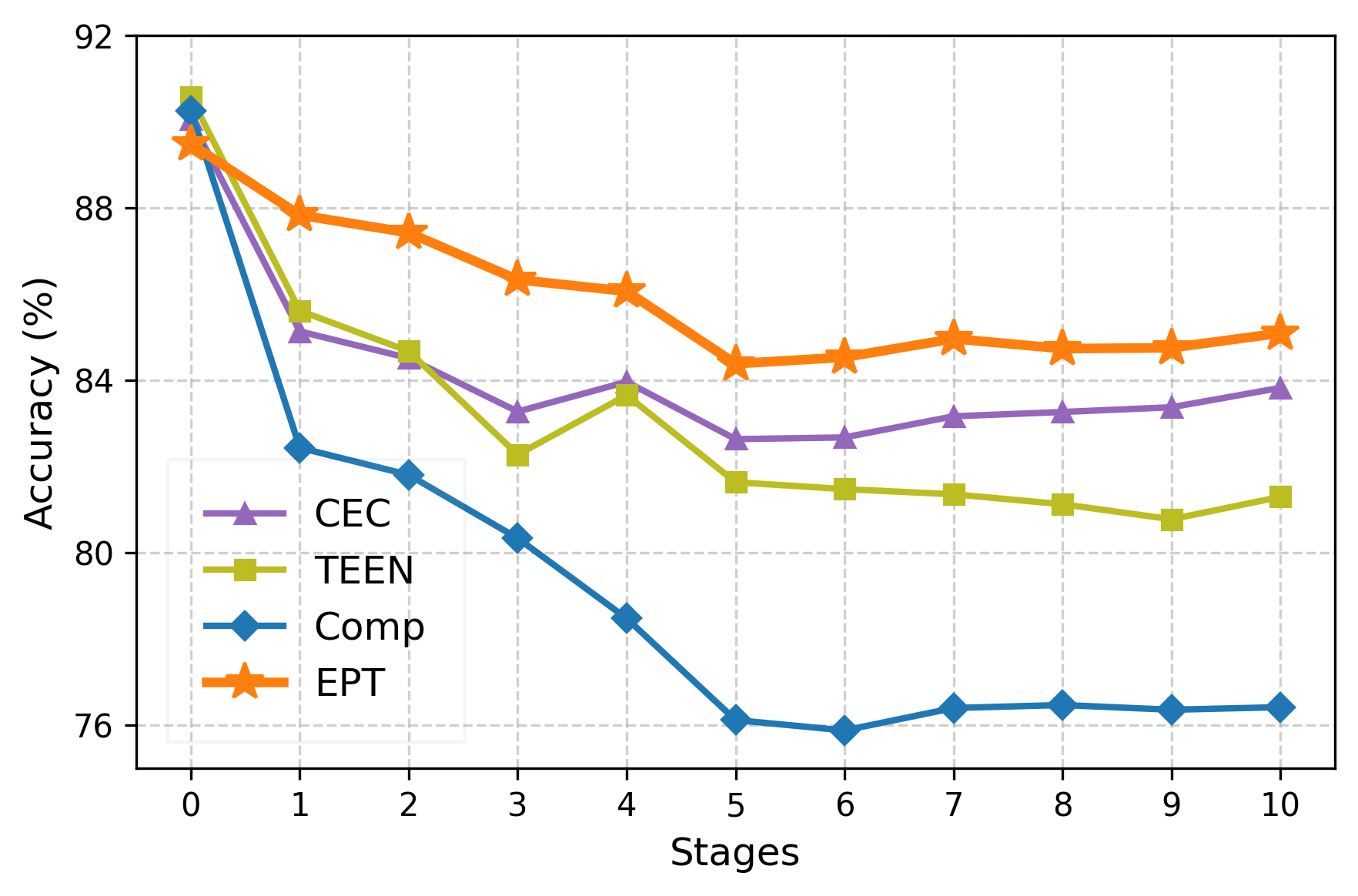}
	\caption{Performance comparison of prototype-based methods using the same pre-trained DINOv2 as feature extractor.}
	\label{fig:cub_dino_diff_method}
\end{figure}

\noindent\textbf{Analysis of different prototype-based methods.} To verify that the proposed fine-tuning strategy can efficiently learn prototypes, we employ the same backbone network and compare it with different previous prototype learning methods. The experimental results are shown in Fig.~\ref{fig:cub_dino_diff_method}. It can be observed that although our method performs slightly worse on the initial base classes compared to certain baseline models, it demonstrates superior and stable performance in each subsequent incremental learning stage. This indicates that our method has stronger feature retention capabilities for learned categories, effectively mitigating catastrophic forgetting. Furthermore, compared to the other methods, our strategy requires fine-tuning only a small number of parameters related to the prototypes, without updating backbones. This approach significantly reduces the risk of overfitting and enhances the representational robustness of the prototypes. These results further demonstrate that the effectiveness of our proposed strategy, rather than relying solely on a powerful pre-trained backbone.

%%%%%%%%%%%%%%%%%%%%%%%%%%%%%%%%%%%%%%%%%%%%%%%%%%%%%%%%%%%%%%%%%%%%%%%%%%%%%%%
%%%%%%%%%%%%%%%%%%%%%%%%%%%%%%%%%%%%%%%%%%%%%%%%%%%%%%%%%%%%%%%%%%%%%%%%%%%%%%%

\end{document}